\pdfoutput=1
\documentclass[11pt,a4paper]{article}
\usepackage[hyperref]{acl2021}
\usepackage{times}
\usepackage{latexsym}
\usepackage{graphicx}
\usepackage{tabularx}
\usepackage{amssymb}
\usepackage{amsmath}
\usepackage{amsthm}
\usepackage{booktabs}
\usepackage{algorithm}
\usepackage{algorithmic}
\usepackage{multirow}

\usepackage{microtype}

\aclfinalcopy 

\title{The HW-TSC's Offline Speech Translation Systems for IWSLT 2021 Evaluation}

\author{Minghan Wang\textsuperscript{\rm 1},
  Yuxia Wang\textsuperscript{\rm 1},
  Chang Su\textsuperscript{\rm 1},
  Jiaxin Guo\textsuperscript{\rm 1},
  Yingtao Zhang\textsuperscript{\rm 1},
  Yujia Liu\textsuperscript{\rm 1},
  \\
  
  {\bf Min Zhang\textsuperscript{\rm 1},}
  {\bf Shimin Tao\textsuperscript{\rm 1},}
  {\bf Xingshan Zeng\textsuperscript{\rm 2},}
  {\bf Liangyou Li\textsuperscript{\rm 2},}
  {\bf Hao Yang\textsuperscript{\rm 1},}
  {\bf Ying Qin\textsuperscript{\rm 1}}
  \\
  \textsuperscript{\rm 1}Huawei Translation Services Center \\
  \textsuperscript{\rm 2}Huawei Noah’s Ark Lab \\
  \tt \{wangminghan,wangyuxia5,suchang8,guojiaxin1,zhangyingtao9,\\
  \tt liuyujia13,zhangmin186,taoshimin,zeng.xingshan,\\
  \tt liliangyou,yanghao30,qinying\}@huawei.com}

\date{}

\begin{document}
\maketitle
\begin{abstract}

This paper describes our work in participation of the IWSLT-2021 offline speech translation task. 
Our system was built in a cascade form, including a speaker diarization module, an Automatic Speech Recognition (ASR) module and a Machine Translation (MT) module. 
We directly use the LIUM SpkDiarization tool as the diarization module. 
The ASR module is trained with three ASR datasets from different sources, by multi-source training, using a modified Transformer encoder.
The MT module is pretrained on the large-scale WMT news translation dataset and fine-tuned on the TED corpus.
Our method achieves 24.6 BLEU score on the 2021 test set.
\end{abstract}

\section{Introduction}
Speech translation (ST) system aims to translate the speech of source language to the text in target language. 
There are two types of ST systems: cascade and end-to-end.
The cascade system consists of several sequentially-connected sub-systems, where they are trained independently, the output of the first module is the input of the second, and so on. 
By contrast, end-to-end systems generate the target text directly from the audio signal, without intermediate outputs, in which all parameters are updated jointly \citep{DBLP:conf/acl/BentivogliCGKMN20}.

In IWSLT-2021 offline speech translation task, participants are welcome to build any one of the two types.
Participants can build their system merely on the provided data from the committee which is considered as constrained track, or, adding external publicly available resources, which is considered as unconstrained. 
The test set is extracted from TED talks --- translation from English speech to German text, being similar to the setup of last few years \citep{anastasopoulos-etal-2021-findings}.

Although end-to-end ST system avoids error propagation, we go with cascade system due to its overwhelming advantage of rich training sources.
Our system is composed of three modules, including the speaker diarization module --- we apply a off-shelf tool LIUM\_SpkDiarization \citep{meignier:hal-01433518}, the ASR module and the MT module which are trained on our own.

The paper is organized as below: Section 2 introduces the dataset applied in training.
Section 3 describes the details of each module.
Experimental setup and results are demonstrated in Section 4 with conclusion in the final Section 5.

\section{Data}
\label{sec2}

\begin{table}[t]
\centering
\begin{tabular}{@{}clcc@{}}
\toprule
\multicolumn{1}{l}{} & Corpora & Size & Time (Hr) \\ \midrule
\multirow{3}{*}{ASR} & LibriSpeech & 281K & 960 \\
 & MuST-C V2 & 248K & 435 \\
 & CoVoST & 564K & 900 \\ \midrule
\multirow{3}{*}{MT} & WMT bitext & 27.7M & - \\
 & Back translated & 521M & - \\
 & TED corpus & 209K & - \\ \bottomrule
\end{tabular}
\caption{Statistical information of corpus applied for ASR and MT training.}
\label{tab_data}
\end{table}

We use the LibriSpeech \citep{DBLP:conf/icassp/PanayotovCPK15}, MuST-C V2 \citep{DBLP:journals/csl/CattoniGBNT21} and CoVoST \citep{DBLP:conf/lrec/WangPWG20} to train the ASR module.
The audio of LibrisSpeech is derived from audio books reading and the text is case-insensitive without punctuation.
MuST-C is a multilingual dataset recorded from the TED-talks, we only use the English data for ASR task.
CoVoST is also a multilingual speech translation dataset based on Common voice, and the content is open-domain.
Both MuST-C and CoVoST have case-sensitive text with punctuation.

To pretrain the MT module, we employ all available bilingual text, as well as the news crawl monolingual text provided in the WMT2019 news translation task \citep{DBLP:conf/wmt/BarraultBCFFGHH19}, followed by the fine-tuning using TED corpus\footnote{https://wit3.fbk.eu/2017-01-c}. 
Table \ref{tab_data} shows the statistics information of the corpus applied.

The development and test sets are provided on the official site including the data ranging from 2010 to 2021.
Despite offering the segmented version, we do not leverage them in our setting since we find the quality of that is not enough to produce good transcripts.

\section{Method}
\subsection{Speaker Diarization}
The audio file provided in the development and test sets are recordings of complete TED speech.
To obtain sentence-level segmentation, LIUM\_SpkDiarization is employed with default parameter configuration, followed by the removal of audio samples that is less than one second. 

\subsection{Automatic Speech Recognition}
Our ASR model is built upon Transformer architecture \citep{DBLP:conf/nips/VaswaniSPUJGKP17}, but in encoder, to better extract features of audio instead of text, we replace the original word embedding layer with two 1-dimensional convolution layers, with kernel size=5 for both \citep{DBLP:journals/corr/abs-1911-08460}.

We find that the test set of IWSLT --- specifically content domain and writing style, is closer to MuST-C compared to another two corpus mentioned in Section \ref{sec2}, but it's relatively in small scale.
So we expect the model to learn common knowledge of ASR that is domain- and style-invariant from distribution-distant LibriSpeech and CovoST, but on the other hand, being able to decode in a specific style learned from MuST-C.
To this end, we propose to train the model with the approach of multi-source training:


\paragraph{Multi-Source Training}
explicitly provides the data source information as a prior condition to the model, so that the distribution in the text space can be modeled separately.
It's commonly used in multilingual translation \citep{DBLP:journals/corr/HaNW16,DBLP:journals/tacl/JohnsonSLKWCTVW17}, where a language tag is used as the first token of input to the decoder, to steer the model to translate in specific language.

In our ASR model, we use [LS], [MC] and [CV] to represent LibriSpeech, MuST-C and CoVoST. 
During training, they are used as the first token to substitute the [BOS] token on samples from specific data source, which provides the model with an explicit signal to generate text in the style as the tag.
Formally, we define $s \in \mathcal{S}$ as the data source tag and integrate it into the original objective:
$$p(y_i|y<i,s,X)$$
\noindent where $X$ are source tokens, $y_i$ is the target token at current step, $y<i$ are previously generated tokens under the autoregressive setting.
In this way, the probability to predict the next token is additionally conditioned on the data source.

During inference, we force the model to decode in the style of MuST-C by feeding [MC] as the initial token, since the test set of IWSLT task has the same data source with MuST-C (TED-talks).
This method successfully expands the size of training set meanwhile prevents the model from decoding confusingly.

\subsection{Machine Translation}
For the machine translation model, we strictly follow \citet{DBLP:conf/wmt/NgYBOAE19} to pretrain the model on the WMT 2019 news translation corpus including bilingual text and back translated data from monolingual text.
Then we fine-tuned on the TED corpus for domain adaptation.

\section{Experiments}

\begin{table*}[t]
\centering
\begin{tabular}{@{}lcccccc@{}}
\toprule
SET & BLEU & TER & BEER & CharacTER & BLEU(ci) & TER(ci) \\ \midrule
dev2010 & 26.00 & 58.85 & 53.22 & 48.52 & 27.56 & 56.42 \\
tst2010 & 26.37 & 58.72 & 52.21 & 51.45 & 27.95 & 56.27 \\
tst2013 & 29.89 & 55.97 & 53.59 & 47.80 & 31.33 & 53.84 \\
tst2014 & 28.03 & 57.56 & 52.98 & 48.78 & 29.17 & 55.61 \\
tst2015 & 23.20 & 74.90 & 50.59 & 51.63 & 24.37 & 72.84 \\
tst2018 & 22.13 & 70.26 & 51.58 & 52.44 & 23.43 & 67.97 \\ \midrule
tst2020 & 25.40 & - & - & - & - & - \\
tst2021 & 20.3/24.6 & - & - & - & - & - \\ \bottomrule
\end{tabular}
\caption{The experimental results of our systems. Results from dev2010 to tst2018 are from our own evaluation. Scores of tst2020 and tst2021 are from the official report, where tst2020 uses TEDRef and tst2021 uses TEDRef (left) and NewRef (right)}
\label{tab_perf}
\end{table*}

\subsection{Setup}
80 dimensional Mel-Filter bank features are extracted from audio files for ASR training corpus.
Sentencepiece \citep{DBLP:conf/emnlp/KudoR18} is utilised for tokenization on ASR texts with a learned vocabulary restricted to 20000 sub-tokens.
For the MT datasets, we apply moses tokenizer \citep{DBLP:conf/acl/KoehnHBCFBCSMZDBCH07} and BPE \citep{DBLP:conf/acl/SennrichHB16a} for tokenization.

ASR model is configured as: $n_{\text{encoder\_layers}}$ = 12, $n_{\text{decoder\_layers}}$ = 6, $n_{\text{heads}}$ = 16, $d_{\text{hidden}}$ = 1024, $d_{\text{FFN}}$ = 4096.
The NMT model has the standard Transformer-big configuration but with $d_{\text{FFN}}$ set to 8192 \citep{DBLP:conf/wmt/NgYBOAE19}.
All models are implemented with \texttt{fairseq} \citep{DBLP:conf/naacl/OttEBFGNGA19}.

During the training of ASR model, we set the batch size to the maximum of 40,000 frames per card.
Inverse sqrt is used for lr scheduling with warm-up steps set to 10,000 and peak lr set as 5e-4.
Adam is used as the optimizer.
The model is trained on 4 V100 GPUs for 50 epochs.
Parameters for last 4 epochs are averaged.
All audio inputs are augmented with spectral augmentation \citep{DBLP:conf/interspeech/ParkCZCZCL19} and are normalized with utterance cepstral mean and variance normalization (CMVN).

The pretraining of NMT model strictly follows the work of \citet{DBLP:conf/wmt/NgYBOAE19}.
We pretrained two models with different parameters randomly initialized.
Both of them are fine-tuned on TED corpus for 10,000 steps with 32,768 tokens per batch on 4 V100 GPUs.
Adam is used for optimizing with learning rate set to 1e-5.
Parameters for last 4 epochs for each model are averaged as the final parameter. 
While decoding, two models are ensembled for generating better translation.

We use the toolkit from the SLT.KIT\footnote{https://github.com/jniehues-kit/SLT.KIT} for evaluation on all development set, which produces metrics including BLEU \citep{DBLP:conf/acl/PapineniRWZ02}, TER \citep{Snover2006ASO}, BEER \citep{DBLP:conf/emnlp/StanojevicS14} and CharacTER \citep{DBLP:conf/wmt/WangPRN16}.

\begin{table}[t]
\centering
\begin{tabular}{lll}
\hline
 & Independent & Hybrid \\ \hline
LS test-clean & 3.32 & 3.57 \\
MC test-COMMON & 17.23 & 15.64 \\
CV test & 31.08 & 29.25 \\ \hline
\end{tabular}
\caption{WER (word error rate) score of the ASR model evaluated on the test set of three datasets, training with or without the hybrid. LS, MC and CV represents for LibriSpeech, MuST-C and CoVoST.}
\label{tab_asr_perf}
\end{table}

\subsection{Results}

As shown in Table \ref{tab_perf}, our system consistently obtains strong results on development and test sets over years, except for the tst2018.
Since tst2018 contains speech from lectures which distributionally deviates from TED talks. leading to lower performance.
For the tst2020 and tst2021, we only report the BLEU score from the official report because references have not been published yet.

\begin{table}[t]
\centering
\begin{tabular}{@{}lll@{}}
\toprule
 & Pretrain & Fine-tune \\ \midrule
dev2010 & 31.0 & 33.1 \\
tst2010 & 32.7 & 35.2 \\
tst2013 & 35.3 & 37.8 \\
tst2014 & 31.3 & 33.6 \\
tst2015 & 33.7 & 36.4 \\
tst2018 & 29.9 & 32.1 \\ \bottomrule
\end{tabular}
\caption{BLEU score of MT model evaluated on the golden segmentation of development set with and without fine-tuning.}
\label{tab_mt_perf}
\end{table}

We further perform ablation study on ASR model, to evaluate the influence of \textit{multi-source training} --- training using the hybrid of different sources with explicit tag.
Specifically, we compare the model trained on three corpora independently, with the model trained by multi-source training under the same architecture, and note that we evaluate on test sets provided by the same source as each independent training dataset rather than the same standard benchmark.

Table \ref{tab_asr_perf} shows that except from the LibriSpeech (LS), the performance of other two test sets improves significantly by using multi-source training.
We speculate that MC and CV mutually augment each other due to their similar data distribution, while hurts LS because of large domain gap.
Concretely, most sentences of MC and CV are oral language, and they are case-sensitive with punctuation. 
However, LS overall has better quality, this may be attributed to its written language style.

We further analyze the NMT model by evaluating on the golden segmentation of all development set.
Table \ref{tab_mt_perf} demonstrates the performance of the NMT model before and after fine-tuning.
The BLEU score is calculated with sacreBLEU \citep{DBLP:conf/wmt/Post18}.
Significant improvement is observed over all dev sets by continuous fine-tuning, this reveals that even if pretraining over large-scale parallel data, fine-tuning on domain data is of great importance.

\section{Conclusion}
In this paper, we report our work in the IWSLT-2021 offline speech translation task. 
Our system is structured in a cascade form, including off-shelf speaker diarization, ASR and MT model based on Transformer. 
The multi-source training for ASR model is demonstrated to be significantly useful to fully leveraging data sampled from different distributions, reflecting to different domain and style in our setting, which not only enlarges the size of training data, but also steers the model to decode in a specific style.
Large scaled WMT news corpora enables MT model to learn domain-invariant translation knowledge adequately during pretraining, and in-domain TED corpus brings additional improvements by fine-tuning.
We will investigate more in terms of end-to-end models, benefiting from both ASR and ST dataset in our future work.



\bibliographystyle{acl_natbib}
\bibliography{acl2021}


\end{document}